\documentclass[sigplan,screen]{acmart}

\AtBeginDocument{%
  \providecommand\BibTeX{{%
    \normalfont B\kern-0.5em{\scshape i\kern-0.25em b}\kern-0.8em\TeX}}}

\begin{document}

\title{L3Cube-MahaSocialNER: A Social Media based Marathi NER Dataset and BERT models}

\author{Harsh Chaudhari}
\authornote{Authors contributed equally to this research}
\email{harshc640@gmail.com}
\affiliation{%
  \institution{Pune Institute of Computer Technology, L3Cube Labs}
  \state{Pune}
  \country{Maharashtra, India}
}
\author{Anuja Patil}
\authornotemark[1]
\email{anujadp4@gmail.com}
\affiliation{%
  \institution{Pune Institute of Computer Technology, L3Cube Labs}
  \state{Pune}
  \country{Maharashtra, India}
}
\author{Dhanashree Lavekar}
\authornotemark[1]
\email{dclavekar@gmail.com}
\affiliation{%
  \institution{Pune Institute of Computer Technology, L3Cube Labs}
  \state{Pune}
  \country{Maharashtra, India}
}
\author{Pranav Khairnar}
\authornotemark[1]
\email{pranavkhairnar016@gmail.com}
\affiliation{%
  \institution{Pune Institute of Computer Technology, L3Cube Labs}
  \state{Pune}
  \country{Maharashtra, India}
}
\author{Raviraj Joshi}
\authornotemark[1]
\email{ravirajoshi@gmail.com}
\affiliation{%
  \institution{Indian Institute of Technology Madras, L3Cube Labs}
  \state{Chennai}
  \country{Tamilnadu, India}
}

\begin{abstract}
 This work introduces the L3Cube-MahaSocialNER dataset, the first and largest social media dataset specifically designed for Named Entity Recognition (NER) in the Marathi language. The dataset comprises 18,000 manually labeled sentences covering eight entity classes, addressing challenges posed by social media data, including non-standard language and informal idioms. Deep learning models, including CNN, LSTM, BiLSTM, and Transformer models, are evaluated on the individual dataset with IOB and non-IOB notations. The results demonstrate the effectiveness of these models in accurately recognizing named entities in Marathi informal text. The L3Cube-MahaSocialNER dataset offers user-centric information extraction and supports real-time applications, providing a valuable resource for public opinion analysis, news, and marketing on social media platforms. We also show that the zero-shot results of the regular NER model are poor on the social NER test set thus highlighting the need for more social NER datasets. The datasets and models are publicly available at \url{https://github.com/l3cube-pune/MarathiNLP}
\end{abstract}


\maketitle

\section{Introduction}
% \footnote{All authors have equal contributions.}
One of the most crucial NLP tasks is Named Entity Recognition (NER). It involves identifying and categorizing named entities in text, such as individuals, groups, and places, among others \cite{3,sabane2023enhancing,5}. NER is significant because it enables information extraction, document organization, question answering, machine translation, improved search engine functionality, and sentiment analysis. It enhances machines' ability to comprehend and interpret language, providing invaluable data for various applications.
Social media websites have witnessed a rapid surge in popularity in recent years, emerging as a primary source of user-generated content. The vast amount of information shared on these platforms presents a valuable resource for various Named Entity Recognition (NER) tasks. Social media data poses specific challenges, including non-standard language, abbreviations, misspellings, and informal idioms, which are often challenging for conventional NER algorithms to handle. To address these issues and facilitate the development of reliable NER models tailored for social media data, the availability of high-quality, domain-specific datasets is essential. While numerous benchmark datasets exist for NER in general domains, they frequently fail to capture the nuances and complexities of social media text.
Simultaneously, most NER research has focused on high-resource languages, underscoring the importance of creating efficient NER models for low-resource languages like Marathi \cite{joshi2022l3cube_mahanlp}. In this research paper, we introduce our comprehensive social media dataset, named L3cube-MahaSocialNER, specifically curated for named entity recognition models in the Marathi language.

The motivation behind this research work is that social media sites are well-known for their informal speech, slang, abbreviations, and spelling and grammar variants. NER models trained on L3cube-MahaSocialNER datasets can better handle these linguistic nuances, improving their accuracy in recognizing named entities in informal text. The key advantage is that they capture emerging named entities, including trends, products, events, and entities. Utilizing social media data offers quick analysis and reaction in areas like public opinion, news, and marketing. Social media platforms are centered around users and their activities, allowing NER models to extract valuable user-centric information, including user mentions, user profiles, and relationships between users. Thus, the L3cube-MahaSocialNER dataset offers user-centric information extraction capabilities. Finally, this dataset can also support real-time applications.
 
The \textit{L3cube-MahaSocialNER} dataset is created internally and manually annotated. It is the largest Marathi NER dataset that is freely available and is annotated using both IOB and non-IOB notation. It has 18,000 manually labeled sentences arranged into eight entity classes. The original sentences have been taken from a \textit{L3CubeMahaSent} Marathi Sentiment Analysis dataset \cite{2}, and the average length of these sentences is 15 words. The entities consist of locations, organizations, people, and numeric quantities like time, measure, and other entities like dates and designations \cite{1}. We have also presented the Dataset statistics in this research paper.
Additionally, we have also presented the results of deep learning models trained on the \textit{L3Cube-MahaSocialNER} dataset, including the Convolutional Neural Network (CNN), Long-Short Time Memory (LSTM), biLSTM, and Transformer models like mBERT \cite{3}, IndicBERT \cite{20}, XLM-RoBERTa \cite{19}, RoBERTa-Marathi \cite{1}, MahaBERT \cite{7}, MahaROBERTa \cite{7}, and MahaALBERT \cite{7}.

The main contributions of this work are as follows.
\begin{itemize}
    \item We highlight the need for Social NER datasets by showing that the zero-shot results of regular NER models are poor on social media text.
    \item We present L3Cube-MahaSocialNER the first social media-based NER dataset for Marathi. It consists of 18k examples tagged with 8 NER labels split into train-valid-test sets. The label set is consistent with the existing non-social MahaNER dataset thus allowing for cross-domain analysis. The dataset and models are available publicly\footnote{\url{https://github.com/l3cube-pune/MarathiNLP}}.
    \item We show that fine-tuning an existing NER model on the Social NER dataset works the best. This shows the importance of transfer learning from existing NER datasets.
\end{itemize}

\section{Related Work}
The phrase "Named Entity Recognition" was first used in the Natural Language Processing (NLP) field in the early 1990s \cite{1}. NER has been the subject of numerous statistics, machine learning, and deep learning methodologies development over time \cite{6}.
Regarding specific studies, \cite{1} have presented L3Cube-MahaNER, the first significant Marathi named entity recognition dataset that adheres to the gold standard. They have discussed the manual annotation procedures that were used. They have experimented with a variety of model architectures, including CNN, LSTM, biLSTM, and transformers like BERT and RoBERTa.
\cite{2} have presented the Marathi Sentiment Analysis Dataset - L3CubeMahaSent. They have calculated results for sentiment prediction in CNN, Bi-LSTM, ULMFiT, mBERT, and IndicBERT models. The CNN model combined with trainable Indic fastText word embeddings gave the best results in the 2-class classification experiments, slightly outperforming IndicBERT.
\cite{3} demonstrates the value of language-specific pre-training by showing superior performance of monolingual Marathi models built on mahaBERT compared to its multilingual counterpart. The research tries to uncover the shortcomings in the existing NER systems, particularly in the Hindi language, since these systems are trained to perform on specific datasets and do not generate results on general datasets.
\cite{7} have introduced a transformer library providing the APIs to support cutting-edge TensorFlow or PyTorch models for complex tasks like phrase comparison, question answering, machine translation, and summarization, among others.
Named Entity Recognition for social media data is challenging because of its inherent noisiness \cite{10}. The approach employed by \cite{10} produces results in both categories: a 41.86\% F1-score for entities and a 40.24\% F1-score for surface forms. The experimental survey focused on Named Entity Recognition in tweets. The issue of degraded tweets was resolved by introducing a rebuilt NLP pipeline, beginning with chunking and culminating in named-entity recognition \cite{11}. Social Media NER was also conducted for the Chinese language, and the findings indicate that the disparity between social media and traditional text in Chinese is significantly wider than that in English-language corpora of a similar size. This suggests that this area may be an attractive subject for further research \cite{12}.
In their study, \cite{15} utilized the multichannel neural network described here to participate in the Novel and Emerging Named Entity Recognition shared challenge at the EMNLP 2017 Workshop on Noisy User-Generated Text (W-NUT). They used this network to recognize emerging named entities in social media messages.
\section{Compilation of Dataset}

\subsection{Dataset Introduction}

The MahaSocialNER dataset is specifically designed for sociolinguistic Marathi Named Entity Recognition tasks. It comprises three primary partitions: training, testing, and validation. The training set consists of 12,114 sentences, containing a total of 361,812 tags. The test set includes 2,250 sentences and 67,449 tags, while the validation set consists of 1,500 sentences and 44,062 tags. These distinctions make the MahaSocialNER dataset a valuable resource for sociolinguistic NER research in the Marathi language.

% \subsection{Sentence Analytics:}
\begin{table}[!ht]
    \centering
    \begin{tabular}{|l|l|l|}
    \hline
        \textbf{Dataset} & \textbf{Sentence Count} & \textbf{Tag Count} \\ \hline
        Train & 12114 & 361812 \\ \hline
        Test & 2250 & 67449 \\ \hline
        Valid & 1500 & 44062 \\ \hline
    \end{tabular}
    \caption{Sentence Analytics.}
\label{table:1}
\end{table}
\subsection{NON IOB Tags:}
\setlength{\tabcolsep}{15pt} 
\begin{table}[!ht]
    \centering
    \begin{tabular}{|l|l|l|l|}
    \hline
        \textbf{Tags} & \textbf{Train} & \textbf{Test} & \textbf{Valid} \\ \hline
        O & 308184 & 57446 & 37412 \\ \hline
        NEP & 15038 & 2863 & 1977 \\ \hline
        NEO & 11751 & 2248 & 1527 \\ \hline
        NEL & 10186 & 1821 & 1157 \\ \hline
        NEM & 6604 & 1181 & 784 \\ \hline
        NED & 5530 & 1026 & 657 \\ \hline
        ED & 3910 & 768 & 480 \\ \hline
        NETI & 609 & 96 & 68 \\ \hline
    \end{tabular}
    \caption{NON IOB Tag counts on train-test-valid split.}
\label{table:2}
\end{table}

The 'O' tag (Other tag) is the most prevalent in all datasets, indicating that it is the most common tag. In the Train dataset, the 'O' tag occurs 308,184 times, which is significantly higher than other tags. The 'NEP' (Person), 'NEO' (Organization), 'NEL' (Location), 'NEM' (Measure), 'NED' (Date), and 'ED' (Designation) tags also appear in substantial numbers, though less frequently than 'O'. The 'NETI' (Time) tag is the least common across all datasets.

\subsection{IOB Tags}
\setlength{\tabcolsep}{15pt} 
\begin{table}[!ht]
    \centering
    \begin{tabular}{|l|l|l|l|}
    \hline
        \textbf{Tags} & \textbf{Train} & \textbf{Test} & \textbf{Valid} \\ \hline
        O & 308185 & 57445 & 37410 \\ \hline
        B-NEL & 8280 & 1594 & 1014 \\ \hline
        B-NEP & 8035 & 1618 & 1086 \\ \hline
        B-NEO & 7066 & 1359 & 946 \\ \hline
        I-NEP & 7003 & 1244 & 891 \\ \hline
        B-NEM & 5249 & 960 & 627 \\ \hline
        I-NEO & 4685 & 889 & 579 \\ \hline
        B-NED & 4293 & 842 & 537 \\ \hline
        B-ED & 3111 & 624 & 397 \\ \hline
        I-NEL & 1906 & 228 & 145 \\ \hline
        I-NEM & 1354 & 221 & 157 \\ \hline
        I-NED & 1237 & 185 & 122 \\ \hline
        I-ED & 799 & 144 & 83 \\ \hline
        B-NETI & 347 & 61 & 46 \\ \hline
        I-NETI & 262 & 35 & 22 \\ \hline
    \end{tabular}
    \caption{IOB Tag counts on train-test-valid split.}
\label{table:3}
\end{table}
The table \ref{table:3} provides counts for different IOB tags in each dataset.
The 'O' tag is also the most common IOB tag, and its count is consistent across all datasets. IOB tags such as 'B-NEL', 'B-NEP', 'B-NEO', 'I-NEP', 'B-NEM', and 'I-NEO' have significant counts but are much lower in frequency compared to 'O.' The 'B-NETI' and 'I-NETI' tags have the lowest occurrence in all datasets. The IOB tagging scheme is typically used for named entity recognition (NER) tasks and is structured to represent the beginning (B-), inside (I-), or outside (O) of named entities.
\section{Experimental Techniques}
\subsection{Model Architecture:}
NLP has been completely transformed by deep learning, which is now the preferred method for many tasks. Deep learning models are very effective for tasks like named entity recognition because they can automatically learn complicated patterns and representations from enormous volumes of data. Deep learning models explored in this work are discussed below.

\subsection{CNN}

A single 1D convolutional layer is used in this particular model, and the word embeddings have a dimension of 300. NER model training frequently includes training these embeddings. The 1D convolutional layer, which utilizes the 'relu' activation function and has 512 filters with a kernel size of 3, receives the embeddings after that. The output of the Conv1D layer is then fed into a dense layer that has the same size as the output layer and uses the 'softmax' activation function.  There are 8 classes, hence the model produces 8 labels for the output. 

\subsection{LSTM}

A single 1D convolutional layer is used in this particular model, and the word embeddings have a dimension of 300. NER model training frequently includes training these embeddings. The 1D convolutional layer, which utilizes the 'relu' activation function and has 512 filters with a kernel size of 3, receives the embeddings after that. The output of the Conv1D layer is then fed into a dense layer that has the same size as the output layer and uses the 'softmax' activation function.  There are 8 classes, hence the model produces 8 labels for the output. The 'rmsprop' optimizer is used during training.

\subsection{BiLSTM}

A BiLSTM layer replaces the single 1D convolutional layer in the CNN model's model architecture, which is comparable to that of BiLSTM. This particular model uses a BiLSTM layer with 512 hidden units and an embedding vector of 300 dimensions. 16 batches are used in all.

\subsection{BERT}

The state-of-the-art language model BERT (Bidirectional Encoder Representations from Transformers), has substantially transformed natural language processing (NLP) \cite{1}. It employs a bidirectional learning strategy to intricately contextualize representations of words and sentences. BERT delivers cutting-edge performance by pre-training on massive volumes of unlabeled text and fine-tuning for specific NLP tasks, such as named entity recognition (NER) \cite{3}. Due to its contextualized representations, transfer learning capabilities, and efficient design, BERT proves to be a flexible and highly effective model for various NER applications.
mBERT (Multilingual BERT), trained on multiple languages, aims to capture and understand context-sensitive word and phrase representations across a variety of languages. Without the need for distinct models for each language, mBERT \cite{8} can efficiently perform tasks across languages by training on a diverse multilingual corpus. MahaBERT \cite{3} is a powerful monolingual BERT model trained on 752 million Marathi tokens, using Marathi linguistic resources like L3Cube-MahaCorpus. MahaRoBERTa and MahaALBERT are models based on the RoBERTa framework (xlm-roberta-base) and ALBERT-based framework, respectively. They have been pre-trained using L3Cube-MahaCorpus and freely accessible Marathi linguistic resources.

Google AI has developed a language model specializing in Indian languages called MuRIL (Multilingual Representations for Indian Languages). MuRIL is built upon the transformer architecture and trained on a substantial amount of text in 16 Indian languages.

\subsection{Zero shot results}
The non-social media model, i.e. the MahaNER model[9] was tested out-of-the-box on the MahaSocialNER dataset and the results (68.28\%) were very poor as compared to the non-social media F1-score (86.80\%). This highlights the need for social media-based NER models as the regular models might not be optimal. The table \ref{table:4} showcase the zero shot results.

\setlength{\tabcolsep}{2pt} 
\begin{table}[!ht]
    \centering
    \small % Reduce font size
    \begin{tabular}{|l|l|l|l|l|}
    \hline
        \textbf{Model} & \textbf{F1 Score} & \textbf{Precision} & \textbf{Recall} & \textbf{Accuracy } \\ \hline
        MahaNER IOB & 57.74 & 55.95 & 59.65 & 93.26  \\[0.25em] \hline
        
        MahaNER NON IOB & 68.28 & 70.19 & 66.46 & 95.67  \\[0.25em] \hline
        
    \end{tabular}
    \caption{Zero shot results on MahaSocialNER test set using BERT models trained on MahaNER (non-social) data set.}
    \label{tab:zeroshot}
\label{table:4}
\end{table}

% \setlength{\tabcolsep}{3pt} 
% \begin{adjustbox}{}

% \small % Reduce font size

%     \begin{tabular}{|l|l|l|l|l|}
    
%     \hline
%         \\[-0.75em]
%         \textbf{Model} & \textbf{F1 Score} & \textbf{Precision} & \textbf{Recall} & \textbf{Accuracy } \\[0.25em] \hline
%         \\[-0.75em]
        
%         % \multicolumn{5}{c}{\textbf{MahaNER Zero Shot Results}} \\[0.25em] \hline
%         % \\[-0.75em]
%         MahaNER IOB & 57.74 & 55.95 & 59.65 & 93.26  \\[0.25em] \hline
%         \\[-0.75em]
%         MahaNER NON IOB & 68.28 & 70.19 & 66.46 & 95.67  \\[0.25em] \hline
%         \\[-0.75em]
%     \end{tabular}
    
%     \caption{Table to test captions and labels.}
% \label{table:1}
% \end{adjustbox}

\section{Results}
In this section, we present a comprehensive analysis of the performance of various models in the context of Named Entity Recognition (NER) for Marathi text. We provide the results in two distinct tagging schemes: IOB (Inside, Outside, Beginning) and Non-IOB, each offering unique insights into the models' effectiveness.

\subsection{IOB Tagging Results}

In the IOB tagging scheme, significant variations in performance are observed among different models. Multilingual BERT variants such as "google/muril-base-cased" (MuRIL), "xlm-roberta-base" (XLM-RoBERTa), and "ai4bharat/indic-bert" (IndicBert) demonstrated competitive performance. MuRIL achieved the highest F1 score of 83.47\%. Similarly, XLM-RoBERTa and IndicBert showcased strong capabilities with F1 scores of 83.31\% and 83.01\%, respectively. The "l3cube-pune/hindi-marathi-dev-bert" (DevBERT) model delivered better results, achieving an F1 score of 83.61\%.

Among monolingual BERT variants, "l3cube-pune/marathi-ner" (MahaNER-BERT) emerged with the highest F1 score of 84.06\%, accompanied by a precision of 83.53 and recall of 84.6, indicating balanced performance. Other monolingual variants, including "l3cube-pune/marathi-roberta" (MahaRoBERTa), "l3cube-pune/marathi-bert-v2" (MahaBERT v2), and "l3cube-pune/marathi-bert" (MahaBERT), also demonstrated competitive results. Additionally, it is important to note that conventional models, namely CNN, LSTM, and BiLSTM, achieved F1 scores ranging from 68.4 to 87.2. 

The MahaNER-BERT is a regular Marathi NER model, further fine-tuning this model on MahaSocialNER yields the best performance. This shows that the transfer of learning from regular NER to social NER is helpful in practice. Out of the models trained from scratch, the MahaRoBERTa model performs the best.

\subsection{Non-IOB Tagging Results}

In the Non-IOB tagging scheme, MahaNER-BERT achieved the highest F1 score of 88.23. Multilingual models such as MuRIL, XLM-RoBERTa, and IndicBert maintained competitive F1 scores. Similar to the IOB results, the vanilla models, including CNN, LSTM, and BiLSTM, established a robust baseline for Non-IOB tagging, achieving F1 scores ranging from 68.4 to 87.4.

Notably, MahaNER-BERT stands out as the top performer, excelling in both IOB and Non-IOB tagging schemes. In IOB tagging, it achieved an F1 score of 84.06, while in the "Non-IOB Results" section, MahaNER-BERT once again boasts the highest F1 score of 88.23. This demonstrates its consistent superiority in recognizing Marathi NER entities, making it a compelling choice for sociolinguistic Marathi NER tasks. These results also mentioned in the following table \ref{table:5} provide valuable insights for selecting the most suitable model for Marathi NER tasks.

\setlength{\tabcolsep}{15pt} 
\begin{table}[!ht]
    \centering
  
\begin{tabular}{|l|l|l|} 

     \multicolumn{3}{c}{\textbf{\large Models Scores on MahaSocialNER Dataset}} \\[0.5em] \hline
        \textbf{Model} & \textbf{IOB} & \textbf{NON IOB} \\[0.25em] \hline

        \textbf{MahaNER-BERT} & \textbf{84.06} & \textbf{88.23} \\[0.25em] \hline
        
        MahaRoBERTa & 83.97 & 86.76 \\[0.25em] \hline
        
        MahaBERT v2 & 83.94 & 86.55 \\[0.25em] \hline
       
        DevRoBERTa & 83.86 & 86.81 \\[0.25em] \hline
      
        DevBERT & 83.61 & 87 \\[0.25em] \hline
     
        MuRIL & 83.47 & 86.44 \\[0.25em] \hline
      
        IndicBert & 83.01 & 86.06 \\[0.25em] \hline
        
        mbert & 82.45 & 85.72 \\[0.25em] \hline
      
        BiLSTM + MahaFT & 80.2 & 87.4 \\[0.25em] \hline

        MahaNER-BERT (zero-shot) & 57.74 & 68.28 \\[0.25em] \hline

          \end{tabular}
\caption{Comparison of different models on the MahaSocialNER test set (F1 Score metric).}
    
\label{table:5}
\end{table}

\section{Conclusion}

In this work, we have introduced the L3Cube-MahaSocialNER dataset. This is, by far, the first and largest social media dataset designed expressly to address the issues presented by social media data used for named entity recognition (NER) in the Marathi language, consisting of 18,000 distinct sentences. We employed IOB and non-IOB notations in deep learning models, including CNN, LSTM, biLSTM, and transformers like BERT, to obtain the results. The highest accuracy was observed for the MahaNER-BERT model followed by MahaRoBERTa. This shows that fine-tuning an existing NER model works better than models trained from scratch. Overall, this NER dataset fills an important gap in the field, providing researchers and practitioners with a valuable resource to further enhance the performance and understanding of NER models in the context of social media text.

\begin{acks}
This work was done under the L3Cube Pune mentorship program. We would like to express our gratitude towards our mentors at L3Cube for their continuous support and encouragement. This work is a part of the L3Cube-MahaNLP project \cite{joshi2022l3cube_mahanlp}.
\end{acks}

\bibliographystyle{ACM-Reference-Format}
\bibliography{main}


\begin{thebibliography}{15}

%%% the Web address.

\ifx \showCODEN    \undefined \def \showCODEN     #1{\unskip}     \fi
\ifx \showDOI      \undefined \def \showDOI       #1{#1}\fi
\ifx \showISBNx    \undefined \def \showISBNx     #1{\unskip}     \fi
\ifx \showISBNxiii \undefined \def \showISBNxiii  #1{\unskip}     \fi
\ifx \showISSN     \undefined \def \showISSN      #1{\unskip}     \fi
\ifx \showLCCN     \undefined \def \showLCCN      #1{\unskip}     \fi
\ifx \shownote     \undefined \def \shownote      #1{#1}          \fi
\ifx \showarticletitle \undefined \def \showarticletitle #1{#1}   \fi
\ifx \showURL      \undefined \def \showURL       {\relax}        \fi
% invisible to TeX
\providecommand\bibfield[2]{#2}
\providecommand\bibinfo[2]{#2}
\providecommand\natexlab[1]{#1}
\providecommand\showeprint[2][]{arXiv:#2}

\bibitem[Aguilar et~al\mbox{.}(2017)]%
        {10}
\bibfield{author}{\bibinfo{person}{Gustavo Aguilar}, \bibinfo{person}{Suraj
  Maharjan}, \bibinfo{person}{Adrian~Pastor L{\'o}pez-Monroy}, {and}
  \bibinfo{person}{Thamar Solorio}.} \bibinfo{year}{2017}\natexlab{}.
\newblock \showarticletitle{A Multi-task Approach for Named Entity Recognition
  in Social Media Data}. In \bibinfo{booktitle}{\emph{Proceedings of the 3rd
  Workshop on Noisy User-generated Text}}. \bibinfo{publisher}{Association for
  Computational Linguistics}, \bibinfo{address}{Copenhagen, Denmark},
  \bibinfo{pages}{148--153}.
\newblock
\urldef\tempurl%
\url{https://doi.org/10.18653/v1/W17-4419}
\showDOI{\tempurl}


\bibitem[Carreras et~al\mbox{.}(2002)]%
        {8}
\bibfield{author}{\bibinfo{person}{Xavier Carreras}, \bibinfo{person}{Lluís
  Màrquez}, {and} \bibinfo{person}{Lluís Padró}.}
  \bibinfo{year}{2002}\natexlab{}.
\newblock \showarticletitle{Named Entity Extraction using AdaBoost}.
  \bibinfo{pages}{1--4}.
\newblock
\urldef\tempurl%
\url{https://doi.org/10.3115/1118853.1118857}
\showDOI{\tempurl}


\bibitem[Joshi(2022a)]%
        {6}
\bibfield{author}{\bibinfo{person}{Raviraj Joshi}.}
  \bibinfo{year}{2022}\natexlab{a}.
\newblock \showarticletitle{L3Cube-HindBERT and DevBERT: Pre-Trained BERT
  Transformer models for Devanagari based Hindi and Marathi Languages}.
\newblock \bibinfo{journal}{\emph{arXiv preprint arXiv:2211.11418}}.
\newblock


\bibitem[Joshi(2022b)]%
        {7}
\bibfield{author}{\bibinfo{person}{Raviraj Joshi}.}
  \bibinfo{year}{2022}\natexlab{b}.
\newblock \showarticletitle{L3Cube-MahaCorpus and MahaBERT: Marathi Monolingual
  Corpus, Marathi BERT Language Models, and Resources}. In
  \bibinfo{booktitle}{\emph{Proceedings of the WILDRE-6 Workshop within the
  13th Language Resources and Evaluation Conference}}.
  \bibinfo{pages}{97--101}.
\newblock


\bibitem[Joshi(2022c)]%
        {joshi2022l3cube_mahanlp}
\bibfield{author}{\bibinfo{person}{Raviraj Joshi}.}
  \bibinfo{year}{2022}\natexlab{c}.
\newblock \showarticletitle{L3cube-mahanlp: Marathi natural language processing
  datasets, models, and library}.
\newblock \bibinfo{journal}{\emph{arXiv preprint arXiv:2205.14728}}
  (\bibinfo{year}{2022}).
\newblock


\bibitem[Jung(2011)]%
        {11}
\bibfield{author}{\bibinfo{person}{Jason Jung}.}
  \bibinfo{year}{2011}\natexlab{}.
\newblock \showarticletitle{Towards Named Entity Recognition Method for
  Microtexts in Online Social Networks: A Case Study of Twitter}.
\newblock \bibinfo{journal}{\emph{Expert Systems with Applications}}
  \bibinfo{volume}{39}, \bibinfo{pages}{563--564}.
\newblock
\urldef\tempurl%
\url{https://doi.org/10.1109/ASONAM.2011.39}
\showDOI{\tempurl}


\bibitem[Kakwani et~al\mbox{.}(2020)]%
        {20}
\bibfield{author}{\bibinfo{person}{Divyanshu Kakwani}, \bibinfo{person}{Anoop
  Kunchukuttan}, \bibinfo{person}{Satish Golla}, \bibinfo{person}{Gokul N.C.},
  \bibinfo{person}{Avik Bhattacharyya}, \bibinfo{person}{Mitesh~M. Khapra},
  {and} \bibinfo{person}{Pratyush Kumar}.} \bibinfo{year}{2020}\natexlab{}.
\newblock \showarticletitle{{I}ndic{NLPS}uite: Monolingual Corpora, Evaluation
  Benchmarks and Pre-trained Multilingual Language Models for {I}ndian
  Languages}. In \bibinfo{booktitle}{\emph{Findings of the Association for
  Computational Linguistics: EMNLP 2020}}. \bibinfo{publisher}{Association for
  Computational Linguistics}, \bibinfo{address}{Online},
  \bibinfo{pages}{4948--4961}.
\newblock
\urldef\tempurl%
\url{https://doi.org/10.18653/v1/2020.findings-emnlp.445}
\showDOI{\tempurl}


\bibitem[Kulkarni et~al\mbox{.}(2021)]%
        {2}
\bibfield{author}{\bibinfo{person}{Atharva Kulkarni}, \bibinfo{person}{Meet
  Mandhane}, \bibinfo{person}{Manali Likhitkar}, \bibinfo{person}{Gayatri
  Kshirsagar}, {and} \bibinfo{person}{Raviraj Joshi}.}
  \bibinfo{year}{2021}\natexlab{}.
\newblock \showarticletitle{{L}3{C}ube{M}aha{S}ent: A {M}arathi Tweet-based
  Sentiment Analysis Dataset}. In \bibinfo{booktitle}{\emph{Proceedings of the
  Eleventh Workshop on Computational Approaches to Subjectivity, Sentiment and
  Social Media Analysis}}. \bibinfo{publisher}{Association for Computational
  Linguistics}, \bibinfo{address}{Online}, \bibinfo{pages}{213--220}.
\newblock
\urldef\tempurl%
\url{https://aclanthology.org/2021.wassa-1.23}
\showURL{%
\tempurl}


\bibitem[Lin et~al\mbox{.}(2017)]%
        {15}
\bibfield{author}{\bibinfo{person}{Bill~Y. Lin}, \bibinfo{person}{Frank Xu},
  \bibinfo{person}{Zhiyi Luo}, {and} \bibinfo{person}{Kenny Zhu}.}
  \bibinfo{year}{2017}\natexlab{}.
\newblock \showarticletitle{Multi-channel {B}i{LSTM}-{CRF} Model for Emerging
  Named Entity Recognition in Social Media}. In
  \bibinfo{booktitle}{\emph{Proceedings of the 3rd Workshop on Noisy
  User-generated Text}}. \bibinfo{publisher}{Association for Computational
  Linguistics}, \bibinfo{address}{Copenhagen, Denmark},
  \bibinfo{pages}{160--165}.
\newblock
\urldef\tempurl%
\url{https://doi.org/10.18653/v1/W17-4421}
\showDOI{\tempurl}


\bibitem[Litake et~al\mbox{.}(2023)]%
        {3}
\bibfield{author}{\bibinfo{person}{Onkar Litake}, \bibinfo{person}{Maithili
  Sabane}, \bibinfo{person}{Parth Patil}, \bibinfo{person}{Aparna Ranade},
  {and} \bibinfo{person}{Raviraj Joshi}.} \bibinfo{year}{2023}\natexlab{}.
\newblock \showarticletitle{Mono versus multilingual bert: A case study in
  hindi and marathi named entity recognition}. In
  \bibinfo{booktitle}{\emph{Proceedings of 3rd International Conference on
  Recent Trends in Machine Learning, IoT, Smart Cities and Applications: ICMISC
  2022}}. Springer, \bibinfo{pages}{607--618}.
\newblock


\bibitem[Litake et~al\mbox{.}(2022)]%
        {1}
\bibfield{author}{\bibinfo{person}{Onkar Litake},
  \bibinfo{person}{Maithili~Ravindra Sabane}, \bibinfo{person}{Parth~Sachin
  Patil}, \bibinfo{person}{Aparna~Abhijeet Ranade}, {and}
  \bibinfo{person}{Raviraj Joshi}.} \bibinfo{year}{2022}\natexlab{}.
\newblock \showarticletitle{{L}3{C}ube-{M}aha{NER}: A {M}arathi Named Entity
  Recognition Dataset and {BERT} models}. In
  \bibinfo{booktitle}{\emph{Proceedings of the WILDRE-6 Workshop within the
  13th Language Resources and Evaluation Conference}}.
  \bibinfo{publisher}{European Language Resources Association},
  \bibinfo{address}{Marseille, France}, \bibinfo{pages}{29--34}.
\newblock
\urldef\tempurl%
\url{https://aclanthology.org/2022.wildre-1.6}
\showURL{%
\tempurl}


\bibitem[Liu et~al\mbox{.}(2019)]%
        {19}
\bibfield{author}{\bibinfo{person}{Yinhan Liu}, \bibinfo{person}{Myle Ott},
  \bibinfo{person}{Naman Goyal}, \bibinfo{person}{Jingfei Du},
  \bibinfo{person}{Mandar Joshi}, \bibinfo{person}{Danqi Chen},
  \bibinfo{person}{Omer Levy}, \bibinfo{person}{Mike Lewis},
  \bibinfo{person}{Luke Zettlemoyer}, {and} \bibinfo{person}{Veselin
  Stoyanov}.} \bibinfo{year}{2019}\natexlab{}.
\newblock \bibinfo{title}{RoBERTa: A Robustly Optimized BERT Pretraining
  Approach}.
\newblock
\newblock
\showeprint[arxiv]{1907.11692}~[cs.CL]


\bibitem[Misal and Haribhakta(2022)]%
        {5}
\bibfield{author}{\bibinfo{person}{Akash Misal} {and}
  \bibinfo{person}{Yashodhara Haribhakta}.} \bibinfo{year}{2022}\natexlab{}.
\newblock \showarticletitle{Transfer Learning for Marathi Named Entity
  Recognition}. In \bibinfo{booktitle}{\emph{2022 4th International Conference
  on Advances in Computing, Communication Control and Networking (ICAC3N)}}.
  \bibinfo{pages}{1487--1491}.
\newblock
\urldef\tempurl%
\url{https://doi.org/10.1109/ICAC3N56670.2022.10074266}
\showDOI{\tempurl}


\bibitem[Peng and Dredze(2015)]%
        {12}
\bibfield{author}{\bibinfo{person}{Nanyun Peng} {and} \bibinfo{person}{Mark
  Dredze}.} \bibinfo{year}{2015}\natexlab{}.
\newblock \showarticletitle{Named Entity Recognition for {C}hinese Social Media
  with Jointly Trained Embeddings}. In \bibinfo{booktitle}{\emph{Proceedings of
  the 2015 Conference on Empirical Methods in Natural Language Processing}}.
  \bibinfo{publisher}{Association for Computational Linguistics},
  \bibinfo{address}{Lisbon, Portugal}, \bibinfo{pages}{548--554}.
\newblock
\urldef\tempurl%
\url{https://doi.org/10.18653/v1/D15-1064}
\showDOI{\tempurl}


\bibitem[Sabane et~al\mbox{.}(2023)]%
        {sabane2023enhancing}
\bibfield{author}{\bibinfo{person}{Maithili Sabane}, \bibinfo{person}{Aparna
  Ranade}, \bibinfo{person}{Onkar Litake}, \bibinfo{person}{Parth Patil},
  \bibinfo{person}{Raviraj Joshi}, {and} \bibinfo{person}{Dipali Kadam}.}
  \bibinfo{year}{2023}\natexlab{}.
\newblock \showarticletitle{Enhancing Low Resource NER using Assisting Language
  and Transfer Learning}. In \bibinfo{booktitle}{\emph{2023 2nd International
  Conference on Applied Artificial Intelligence and Computing (ICAAIC)}}. IEEE,
  \bibinfo{pages}{1666--1671}.
\newblock


\end{thebibliography}

\vspace{5cm}
\appendix
\section{Appendix: }
% tried adding the Vanilla code here but wasn't working
\subsection{Model Repository}
\begin{table}[!ht]
    \centering
    \setlength{\tabcolsep}{20pt} 
    \small
    \begin{tabular}{|l|l|p{8cm}|}
    \hline
        \textbf{Model Name} & \textbf{URL} & \textbf{Description}\\ \hline
        MahaSocialNER-BERT & \href{https://huggingface.co/l3cube-pune/marathi-social-ner}{marathi-social-ner} & base model -- MahaBERT, training data -- MahaSocialNER in non-IOB format.\\ \hline
        MahaSocialNER-BERT-IOB & \href{https://huggingface.co/l3cube-pune/marathi-social-ner-iob}{marathi-social-ner-iob} & base model -- MahaBERT, training data -- MahaSocialNER in IOB format. \\ \hline
        MahaNER-Mixed-BERT & \href{https://huggingface.co/l3cube-pune/marathi-mixed-ner}{marathi-mixed-ner} & base model -- DevBERT, training data -- (MahaNER + MahaSocialNER) in non-IOB format.\\ \hline
        MahaNER-Mixed-BERT-IOB & \href{https://huggingface.co/l3cube-pune/marathi-mixed-ner-iob}{marathi-mixed-ner-iob} & base model -- DevBERT, training data -- (MahaNER + MahaSocialNER) dataset in IOB format.\\ \hline
    \end{tabular}
    \caption{Model Repository.}
\label{table:6}
\end{table}

\subsection{Vanilla Code - NON IOB:}

\begin{table*}[!ht]
    \centering
    \setlength{\tabcolsep}{45pt} 
    \small
    \begin{tabular}{|l|l|l|l|}
    \hline
        \textbf{Model} & \textbf{F1 Score} & \textbf{Precision} & \textbf{Recall}\\ \hline
        \multicolumn{4}{|c|}{Random Embeddings Trainable True} \\[0.25em] \hline
        CNN & 83.8 & 87 & 81.1\\[0.25em] \hline
        LSTM & 78.5 & 83.9 & 75.1\\[0.25em] \hline
        BiLSTM & 79 & 88.1 & 72.7\\[0.25em] \hline
        \multicolumn{4}{|c|}{MahaFT Fast Text Trainable False} \\[0.25em] \hline    
        CNN & 86.4 & 89.2 & 84.1\\[0.25em] \hline       
        LSTM & 84.2 & 85.7 & 83.4\\[0.25em] \hline     
        BiLSTM & 87.2 & 87.7 & 86.8\\[0.25em] \hline   
        \multicolumn{4}{|c|}{MahaFT Fast Text Trainable True}\\[0.25em] \hline  
        CNN & 86.9 & 88.6 & 85.6\\[0.25em] \hline       
        LSTM & 84.7 & 84.4 & 85.4\\[0.25em] \hline  
        \textbf{BiLSTM} & \textbf{87.4} & \textbf{87.1} & \textbf{87.8}\\[0.25em] \hline
    \end{tabular}
    \caption{Vanilla Code - NON IOB.}
\label{table:7}
\end{table*}

\clearpage

\subsection{Vanilla Code - IOB Tags:}

\begin{table}[!ht]
    \centering
    \setlength{\tabcolsep}{45pt} 
    \small
    \begin{tabular}{|l|l|l|l|}
    \hline
       \textbf{Model} & \textbf{F1 Score} & \textbf{Precision} & \textbf{Recall} \\[0.25em] \hline
       \multicolumn{4}{|c|}{Random Embeddings Trainable True} \\[0.25em] \hline 
        CNN & 75.9 & 78.7 & 74\\[0.25em] \hline     
        LSTM & 68.4 & 76 & 64.1\\[0.25em] \hline        
        BiLSTM & 68.9 & 80.8 & 62.2\\[0.25em] \hline    
        \multicolumn{4}{|c|}{MahaFT Fast Text Trainable False} \\[0.25em] \hline  
        CNN & 79.2 & 79.4 & 79.8\\[0.25em] \hline        
        LSTM & 77.2 & 76.3 & 79.8\\[0.25em] \hline     
        \textbf{BiLSTM} & \textbf{80.3} & \textbf{82.3} & \textbf{79.3}\\[0.25em] \hline
        \multicolumn{4}{|c|}{MahaFT Fast Text Trainable True}\\[0.25em] \hline
        CNN & 79.8 & 80.8 & 80\\[0.25em] \hline
        LSTM & 78.5 & 79.5 & 79.2\\[0.25em] \hline  
        BiLSTM & 80.2 & 81.8 & 79.2\\[0.25em] \hline   
    \end{tabular}
    \caption{Vanilla Code - IOB.}
\label{table:8}
\end{table}

% \setlength{\tabcolsep}{15pt} 
% % \begin{adjustbox}{width=350,center}
% \begin{adjustbox}{}
% \small % Reduce font size
%     \begin{tabular}{l l l l}
%     \hline
%         \\[-0.75em]
%         \textbf{Model} & \textbf{F1 Score} & \textbf{Precision} & \textbf{Recall} \\[0.25em] \hline
%         \\[-0.75em]
        
%         \multicolumn{4}{c}{\Random Embeddings Trainable True} \\[0.25em] \hline
%         \\[-0.75em]
%         CNN & 75.9 & 78.7 & 74\\[0.25em] \hline
%         \\[-0.75em]
%         LSTM & 68.4 & 76 & 64.1\\[0.25em] \hline
%         \\[-0.75em]
%         BiLSTM & 68.9 & 80.8 & 62.2\\[0.25em] \hline
%         \\[-0.75em]

%         \multicolumn{4}{c}{\MahaFT Fast Text Trainable False} \\[0.25em] \hline
%         \\[-0.75em]
%         CNN & 79.2 & 79.4 & 79.8\\[0.25em] \hline
%         \\[-0.75em]
%         LSTM & 77.2 & 76.3 & 79.8\\[0.25em] \hline
%         \\[-0.75em]
%         \textbf{BiLSTM} & \textbf{80.3} & \textbf{82.3} & \textbf{79.3}\\[0.25em] \hline
%         \\[-0.75em]

%         \multicolumn{4}{c}{MahaFT Fast Text Trainable True}\\[0.25em] \hline
%         \\[-0.75em]
%         CNN & 79.8 & 80.8 & 80\\[0.25em] \hline
%         \\[-0.75em]
%         LSTM & 78.5 & 79.5 & 79.2\\[0.25em] \hline
%         \\[-0.75em]
%         BiLSTM & 80.2 & 81.8 & 79.2\\[0.25em] \hline
%         \\[-0.75em]
%     \end{tabular}
% \end{adjustbox}

\clearpage
\subsection{BERT models on IOB and NON-IOB dataset }

\begin{table}[htb]
  \centering

% \subsection{IOB}
% IOB begin
% \setlength{\tabcolsep}{15pt} 
\setlength{\tabcolsep}{15pt}
 \small % Reduce font size
% \begin{adjustbox}{width=300,center}
    \begin{tabular}{|l|l|l|l|l|}    
    
        \multicolumn{5}{c}{\textbf{\large BERT models on IOB Dataset}} \\[0.5em] \hline
        \textbf{Model} & \textbf{F1 Score} & \textbf{Precision} & \textbf{Recall} & \textbf{Accuracy } \\[0.25em] \hline

        \multicolumn{5}{|c|}{{Multilingual BERT Variants}} \\[0.25em] \hline
        
        google/muril-base-cased (MuRIL) & 83.47 & 82.46 & 84.51 & 96.76  \\[0.25em] \hline
        
        xlm-roberta-base (XLM-RoBERTa) & 83.31 & 82.91 & 83.72 & 96.65  \\[0.25em] \hline
    
        ai4bharat/indic-bert (IndicBert) & 83.01 & 83.38 & 82.64 & 96.65  \\[0.25em] \hline
    
        bert-base-multilingual-cased (mbert) & 82.45 & 81.93 & 82.98 & 96.63  \\[0.25em] \hline

        \multicolumn{5}{|c|}{{Monolingual BERT Variants}} \\[0.25em] \hline
 
        \textbf{l3cube-pune/marathi-ner (MahaNER-BERT)} & \textbf{84.06} & \textbf{83.53} & \textbf{84.6} & \textbf{96.96} \\[0.25em] \hline

        l3cube-pune/marathi-roberta (MahaRoBERTa) & 83.97 & 83.43 & 84.51 & 96.76  \\[0.25em] \hline
     
        l3cube-pune/marathi-bert-v2 (MahaBERT v2) & 83.94 & 83.45 & 84.43 & 96.72  \\[0.25em] \hline
      
        l3cube-pune/marathi-albert-v2 (MahaAlBERT) & 83.17 & 84.01 & 82.35 & 96.51  \\[0.25em] \hline
    
        l3cube-pune/marathi-bert (MahaBERT) & 80.33 & 78.88 & 81.85 & 96.2  \\[0.25em] \hline
           
        l3cube-pune/marathi-roberta (MahaRoBERTa) & 79.68 & 79.12 & 80.26 & 95.99  \\[0.25em] \hline
       
        flax-community/roberta-hindi (RoBERTa Hindi) & 74.16 & 73.89 & 74.43 & 94.93  \\[0.25em] \hline

        \multicolumn{5}{|c|}{{Bilingual BERT Variants}} \\[0.25em] \hline
      
        %\centerline{Bilingual BERT Variants} \\ \hline
        l3cube-pune/hindi-marathi-dev-roberta (DevRoBERTa) & 83.86 & 82.98 & 84.75 & 96.88  \\[0.25em] \hline
     
        l3cube-pune/hindi-marathi-dev-bert (DevBERT) & 83.61 & 82.75 & 84.49 & 96.78  \\[0.25em] \hline
     
        l3cube-pune/hindi-marathi-dev-albert (DevAlBERT) & 83.09 & 83.85 & 82.33 & 96.61  \\[0.25em] \hline
        
        neuralspace-reverie/indic-transformers-hi-roberta & 70.15 & 69.31 & 71.01 & 94.28  \\[0.25em] \hline 

        \end{tabular}
        \newline
        \begin{tabular}{|l|l|l|l|l|} 
     
        \multicolumn{5}{|c|}{\textbf{\large BERT models on NON IOB Dataset}} \\[0.5em] \hline

      \textbf{Model} & \textbf{F1 Score} & \textbf{Precision} & \textbf{Recall} & \textbf{Accuracy } \\[0.25em] \hline

      \multicolumn{5}{|c|}{{Multilingual BERT Variants}} \\[0.25em] \hline
        bert-base-multilingual-cased (mbert) & 85.72 & 83.95 & 87.57 & 97.2  \\[0.25em] \hline
 
        xlm-roberta-base (XLM-RoBERTa) & 86.48 & 84.62 & 88.43 & 97.42  \\[0.25em] \hline

        google/muril-base-cased (MuRIL) & 86.44 & 85.35 & 87.57 & 97.38  \\[0.25em] \hline
   
        ai4bharat/indic-bert (IndicBert) & 86.06 & 85.9 & 86.21 & 97.33  \\[0.25em] \hline

        \multicolumn{5}{|c|}{{Monolingual BERT Variants}} \\[0.25em] \hline
     
        \textbf{l3cube-pune/marathi-ner (MahaNER-BERT)} & \textbf{88.23} & \textbf{86.28} & \textbf{90.28} & \textbf{97.73} \\[0.25em] \hline
   
        l3cube-pune/marathi-bert (MahaBERT) & 86.82 & 85.24 & 88.47 & 97.51  \\[0.25em] \hline
  
        l3cube-pune/marathi-roberta (MahaRoBERTa) & 86.76 & 85.18 & 88.4 & 97.4  \\[0.25em] \hline
      
        l3cube-pune/marathi-bert-v2 (MahaBERT v2) & 86.55 & 85.2 & 87.94 & 97.37  \\[0.25em] \hline
       
        l3cube-pune/marathi-albert-v2 (MahaAlBERT) & 86.34 & 86.01 & 86.68 & 97.26  \\[0.25em] \hline
     
        l3cube-pune/marathi-roberta (MahaRoBERTa) & 82.53 & 81.51 & 83.57 & 96.98  \\[0.25em] \hline

        flax-community/roberta-hindi (RoBERTa Hindi) & 82.25 & 81.42 & 83.1 & 96.58  \\[0.25em] \hline

        \multicolumn{5}{|c|}{{Bilingual BERT Variants}} \\[0.25em] \hline
        
        l3cube-pune/hindi-marathi-dev-bert (DevBERT) & 87 & 85.77 & 88.27 & 97.38  \\[0.25em] \hline
        
        l3cube-pune/hindi-marathi-dev-roberta (DevRoBERTa) & 86.81 & 85.32 & 88.35 & 97.44  \\[0.25em] \hline
   
        l3cube-pune/hindi-marathi-dev-albert (DevAlBERT) & 84.77 & 84.3 & 85.25 & 97.32  \\[0.25em] \hline
      
        neuralspace-reverie/indic-transformers-hi-roberta & 80.47 & 79.01 & 81.98 & 96.41  \\[0.25em] \hline
       
    \end{tabular}

% \end{adjustbox}
\caption{BERT models on IOB and NON-IOB dataset.}
\label{table:8}
% IOB End
\end{table}

\end{document}